\UseRawInputEncoding
\documentclass[letterpaper, 10 pt, conference]{ieeeconf}

\IEEEoverridecommandlockouts
\usepackage{amsmath,amssymb}
\usepackage{graphicx}
\usepackage{placeins}
\usepackage{stfloats}
\usepackage{tikz}
\usepackage{url}
\usetikzlibrary{arrows.meta,positioning,fit,calc,backgrounds,shapes.geometric}

\newcommand{\method}{OPAL}
\newcommand{\methodfull}{Omnidirectional Path-efficient Aerial 3D expLoration}
\newcommand{\nf}{\mbox{\method{}-NF}}

\newcommand{\methodnfp}{\mbox{\method{}-NFP}}
\newcommand{\methodnfpr}[1]{\mbox{\method{}-NFP$_{#1}$}}

\newcommand{\methodv}{\method{}-V}
\newcommand{\methodof}{\method{}-OF}
\newcommand{\methodl}{\method{}-L}

\title{\LARGE \bf
\method{}: \methodfull{} 
}

\author{Yoga Satwik Chappidi$^{1}$ and Avideh Zakhor$^{1}$%
\thanks{$^{1}$Department of Electrical Engineering and Computer Sciences, University of California, Berkeley, Berkeley, CA, USA.}%
}

\begin{document}

\specialpapernotice{This work has been submitted to the IEEE for possible publication. Copyright may be transferred without notice, after which this version may no longer be accessible.}
\maketitle
\thispagestyle{empty}
\pagestyle{empty}

\begin{abstract}

Autonomous exploration is critical for robots mapping unknown environments. Desirable characteristics 
of exploration algorithms include compute efficiency and small traversed distance during the exploration 
process. Motivated by these, we present \methodfull{} (\method{}), an exploration framework centered on deliberate $360^\circ$ yaw rotation at ambiguous branch points rather than compute-heavy global tour planning.
We devise multiple variants of OPAL to determine the frontier-selection strategy once the yaw pan is
completed. One variant is model-free, while others use large language models (LLMs) or vision-language
models (VLMs). We characterize the performance of these variants while varying the vicinity search radius
to include frontiers in the selection process. Through simulations we find that although the time-consuming in-place yaw
rotation increases total exploration time relative to more computationally complex baselines such
as EDEN and FALCON, OPAL is computationally simpler and achieves shorter travel distances and higher
coverage-versus-distance area under the curve. We also show that adjusting the frontier-selection search
radius enables a tradeoff between travel distance and total exploration time. We verify our results on 
a Modal AI drone in two indoor environments by comparing OPAL against FALCON, and 
find that the traveled distance for a variant of OPAL to be as much as 25\% lower than FALCON.

\end{abstract}

\section{Introduction}

Autonomous exploration is a core requirement for robots operating in previously unseen indoor
environments. Systems for inspection, search-and-rescue, and remote documentation all benefit
from the ability to map an unknown building efficiently rather than relying on a map constructed
in advance. Frontier exploration remains attractive because it is reactive and robust
\cite{yamauchi1997frontier}, but the central challenge is not simply detecting new frontiers.
Rather, it is choosing among locally feasible continuations whose downstream value may differ sharply.
In cluttered indoor layouts, a geometrically easy-to-reach frontier can still be a poor global choice
if it leads into a short dead end while a slightly farther branch might open up a much larger region.

Recent exploration systems such as FALCON \cite{zhang2024falcon} and EDEN \cite{dong2025eden} show that
strong exploration performance for time-critical missions is possible with richer global guidance
and heavier planning machinery. Recent omnidirectional systems such as Lantern-Explorer
\cite{zhu2025lantern} likewise show that broad
sensing can support high-speed continuous flight, pairing a $360^\circ$ LiDAR with FUEL-360
frontier generation, dual-nested Traveling Salesman Problem (TSP) viewpoint ordering, and
yaw-aware control to preserve mapping quality during aggressive motion.

This paper asks whether competitive, complete 3D mapping benefits from using wide-field
perception to simplify branch decisions. We propose \methodfull{} (\method{}), designed to study
that question directly. Rather than replacing the underlying
geometric stack, \method{} inserts a lightweight decision layer between frontier generation and
frontier commitment. It can restrict candidate selection to a local reachable set, relax that
constraint with a larger vicinity radius, or remove the vicinity rule entirely while keeping
mapping, planning, and recovery fixed. It is omnidirectional because it can pause at ambiguous
branch states for a fresh in-place $360^\circ$ pan before choosing a continuation. This same
interface also allows us to test multimodal language models explicitly at ambiguous branch states.
Specifically, the model takes as input the pan views and frontier metadata and chooses the next
local continuation.
This frontier-arbitration
framing is inspired by prior work on high-level VLM-guided frontier selection in
\cite{aitha2025vlmthesis,aitha2026cvprw} for 2D exploration with hexapods, but here the emphasis is on
an analysis of pan refresh and reachable-vicinity constraints, ranging from tightly local to
effectively global, within a 3D aerial exploration stack.

The main observation of this paper is that a brief $360^\circ$
pan at genuine branch points and a less restrictive reachable-vicinity rule can reduce computation time, travel distance,
and improve coverage-versus-distance AUC by helping the robot commit to more promising continuations earlier. These improvements come at the expense
of longer travel times due to the time-consuming $360^\circ$ rotation of the drone, which can in principle
be mitigated by mounting the camera depth sensor on a spinning mechanism.

The rest of this paper is organized as follows. Section~\ref{sec:related}
reviews prior work on frontier exploration, global guidance, and omnidirectional sensing.
Section~\ref{sec:method} presents \method{}, including the reachable-vicinity filter, the decision
gate and pan refresh mechanism, the local selector policies, and the recovery logic.
Section~\ref{sec:eval} describes the experimental setup and reports benchmark comparisons,
language-model versus no-model results, structural ablations over locality and vicinity radius,
timing analysis, and hardware validation. Section~\ref{sec:conclusion} includes conclusions and
future work.

\section{Related Work}
\label{sec:related}

Classical frontier exploration remains the backbone of many mapping systems because it offers a
simple interface between known free space and unknown space \cite{yamauchi1997frontier}.
Recent geometric exploration methods extend that formulation with richer viewpoint planning and
more explicit global guidance \cite{bircher2016nbv,selin2019aeplan,kompis2021informed,batinovic2022shadowcasting}.
Receding-horizon and hierarchical aerial systems such as FUEL \cite{zhou2021fuel},
FALCON \cite{zhang2024falcon}, EDEN \cite{dong2025eden}, FSMP \cite{zhang2025fsmp},
FLARE \cite{liu2025flare}, and PTplanner \cite{zhao2026ptplanner} now pursue efficient
large-scale coverage through stronger frontier ordering, unknown-region reasoning, and
global-local planning. Across that broader planner literature, representative strategies include
receding-horizon path planning for exploration and inspection \cite{bircher2018rhp},
incrementally built topological-map guidance \cite{wang2020topological},
graph-based topological exploration \cite{yang2021graph},
multi-resolution frontier planners \cite{batinovic2021multires},
hierarchical global-local frameworks such as TARE \cite{cao2021tare},
dynamic-environment sampling or obstacle-aware planners \cite{xu2021dep,wiman2024daep},
limited Field-of-View (FoV) fast exploration strategies \cite{zhao2023faep},
collector-style frontier management \cite{changoluisa2024collector},
low-memory large-scene planners \cite{huang2024lowmemory},
pathwise information-gain planners \cite{baek2025pipe},
specialized aerial planners for underground or cluttered scenes \cite{rubiosierra2020mines,lu2020rgbd},
surface-driven next-best-view exploration \cite{hardouin2020surfacedriven}, and
prior-topology-guided large-scene exploration \cite{cao2024priortopo}.

A related but distinct line of work uses wide-field sensing to preserve high-speed continuous
exploration. Lantern-Explorer \cite{zhu2025lantern} couples omnidirectional LiDAR with the FUEL-360 exploration
algorithm, combining LiDAR-FOV-aware frontier generation, dual-nested TSP viewpoint ordering,
and yaw-aware control to reduce backtracking and limit destabilizing yaw changes during flight.
\method{} shares the premise that wider fields of view at ambiguous branch points can materially improve exploration, but it uses a finite-FoV depth sensor and a deliberate in-place drone rotation of $360^\circ$ to
achieve broad visual coverage.

At the same time, semantic and language-grounded robotics has motivated a different line of
work in which high-level reasoning guides embodied decisions. Prior work has studied semantic
frontier reasoning \cite{gomez2019semanticfrontier,milas2023asep} and more general multimodal robotic
reasoning \cite{ahn2022saycan,driess2023palme}. Most relevant to the present paper is the line of work represented by
\cite{aitha2025vlmthesis,aitha2026cvprw}, which studies 2D exploration with hexapods. It uses
high-level VLM guidance to select frontiers within a conventional exploration pipeline, using top-down occupancy maps
and frontier-facing RGB images as context. The present paper deals with 3D aerial exploration rather than 2D terrestrial exploration
and carries out a complete characterization of frontier-selection strategies at branch states
following a $360^\circ$ pan.

\section{Method}
\label{sec:method}

\method{} is closely related to FALCON \cite{zhang2024falcon} and shares the same base exploration stack, including the
mapper, planner, and controller. Its main difference is how the robot
chooses its next frontier.

\begin{figure*}[!t]
\centering
\includegraphics[width=0.98\textwidth]{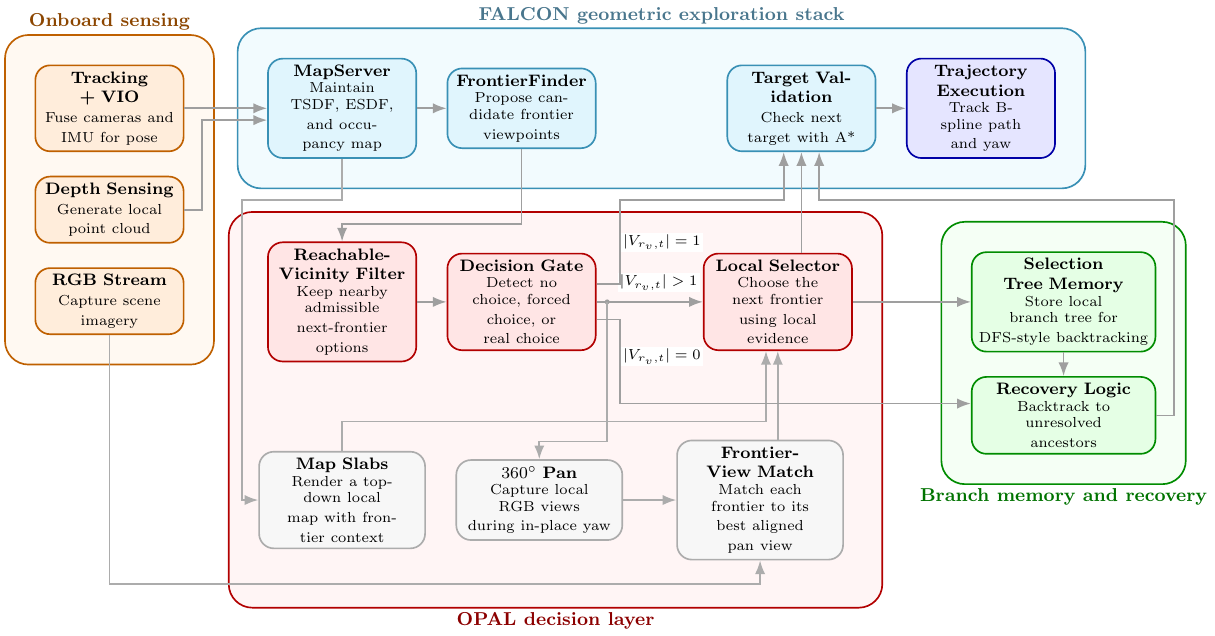}
\caption{Overview of \method{} with exploration stack inherited from FALCON. The orange group is the
onboard sensing layer, the cyan group is the base geometric exploration stack, the pink group is
the \method{} local frontier-choice layer, and the green group stores branch history and handles
recovery using a lightweight search-tree memory inspired by classical depth-first backtracking. The gray blocks are used for collecting optional local-selection inputs, including the map slab,
pan views, and frontier-view matching.}
\label{fig:horus_falcon_stack}
\end{figure*}

Figure~\ref{fig:horus_falcon_stack} shows the overall block diagram of \method{}. The orange block
is the onboard sensing layer. It captures the robot pose, depth measurements, and RGB images that
are used by the rest of the system. The cyan block is the base geometric exploration stack
inherited from FALCON. It uses the sensor inputs to build the map, then generates frontier
viewpoints, checks whether a selected target is valid, and executes the motion command. The pink
middle row, unique to \method{}, consists of the Reachable-Vicinity Filter, Decision Gate, and
Local Selector. The gray blocks below it are optional evidence data used by selected variants and
ablations. The green block, also unique to \method{}, maintains a lightweight history of branch
choices and enables Depth First Search (DFS)-style backtracking when the robot cannot make
progress or exhausts a local region.

The overall pipeline in Fig.~\ref{fig:horus_falcon_stack} works as follows. The onboard sensing
layer first captures the current robot pose, depth measurements, and RGB observations. MapServer
uses the pose and depth inputs to maintain the current voxel map. FrontierFinder then reads that
map and proposes candidate frontier viewpoints. The Reachable-Vicinity Filter keeps only the
frontiers that are both within a given vicinity radius $r_v$ and admissible as the next move from
the current robot pose. The result is $V_{r_v,t}$, a reachable-vicinity set at continuous
planning time $t$.

The Decision Gate determines whether a valid choice is present. If $|V_{r_v,t}|=0$, there is no admissible
local frontier, and the robot hands control to Recovery Logic. If $|V_{r_v,t}|=1$, there is only one
reasonable continuation, and that frontier is passed directly to Target Validation followed by
Trajectory Execution. If $|V_{r_v,t}|>1$, the robot is at a branch point, with several
frontiers as possible next moves, and the system must decide among them.

At such branch points \method{} collects additional evidence by first performing a
$360^\circ$ yaw rotation that captures RGB views before the robot moves. The
Frontier-View Match module then pairs each candidate frontier with the pan image whose viewing
direction is most aligned with that frontier, producing the frontier-aligned images. % illustrated in Fig.~\ref{fig:horus_decision_point_example}(b). 
Another evidence collected is the Map Slab, shown in Fig.~\ref{fig:horus_decision_point_example}, a
top-down image of the local map around the robot showing free space, obstacles, the recent robot
trace, the current pose, candidate frontiers, and decision-point status markers.
The Local Selector uses the matched frontier images, map slab, and frontier metadata such as
coordinates and planner cost to choose the next frontier.
We consider both model-free and language- or vision-model-based variants of OPAL for use in the Local Selector.
In the model-based variants, the local evidence is passed to the model so it can choose among ambiguous
branch continuations.
Regardless of how a frontier is selected, it is stored in
Selection Tree Memory so that Recovery Logic can later revisit earlier alternatives if local
progress is exhausted. The selected frontier is then passed to
Target Validation and Trajectory Execution, which are the same as in FALCON. 

\subsection{Reachable-Vicinity Filter}

The first step in the \method{} decision layer, shown in pink on Fig.~\ref{fig:horus_falcon_stack}, is to reduce the frontier list to the candidates
that influence the robot's immediate next move. This Reachable-Vicinity
Filter is the handoff between FrontierFinder and the \method{} decision layer. 
Let $t \in \mathbb{R}_{\ge 0}$ denote continuous planning time. Let $F_t$ denote the active frontier viewpoints at time $t$. Let $p_t$ be the current
robot pose with translational component $x_t \in \mathbb{R}^3$, and let $a_f \in \mathbb{R}^3$ be
the frontier-average position associated with candidate $f$. We form a reachable-vicinity set
\begin{equation}
V_{r_v,t} = \left\{ f \in F_t \;\middle|\; \|a_f - x_t\| \le r_v,\ q_t(f) = 1 \right\},
\label{eqn:reachable_vicinity_filter}
\end{equation}
where $r_v$ is the vicinity radius and $q_t(f) \in \{0,1\}$ is a binary indicator of whether
frontier $f$ is admissible as the next move. Specifically, $q_t(f)=1$ means that the frontier
viewpoint is a valid next target, meaning it has sufficient clearance from obstacles and
that the planner can find a feasible path to it from the current robot pose. Conversely,
$q_t(f)=0$ means that the frontier viewpoint is not a valid next target. In plain terms, $V_{r_v,t}$ is the set of
frontiers that the robot could reasonably choose next from its current pose and as such it is passed to the Decision Gate.

The value of $r_v$ in Eqn.~\ref{eqn:reachable_vicinity_filter} controls how locally constrained the next decision is. Small values force the
robot to exhaust nearby frontiers before considering broader continuations. Large values relax
that behavior, and the $r_v\to\infty$ limit recovers a globally free frontier choice.

\subsection{Decision Gate for Local Selection}

Once the reachable-vicinity set $V_{r_v,t}$ has been formed, we first determine whether the
Local Selector should be invoked at all. This decision gate is shared across selector policies,
including both the greedy no-model selector and the language-model variants. We invoke the Local
Selector only at ambiguous branch points where several admissible next frontiers compete. Specifically, we treat the cardinality of $V_{r_v,t}$ as the decision gate. If $|V_{r_v,t}|=0$, local progress is
exhausted and recovery must take over. If $|V_{r_v,t}|=1$, typically occurring in corridors, the next move is effectively determined and there is no need for an extra
$360^\circ$ rotation.
Only when $|V_{r_v,t}|>1$ does the system face an ambiguous branch-selection problem. In that case, control
passes to the Local Selector, whose policy may be greedy or model-based depending on the variant.

\subsection{Map Slabs and Frontier-View Match}

\begin{figure}[t]
\centering
\includegraphics[width=0.95\columnwidth]{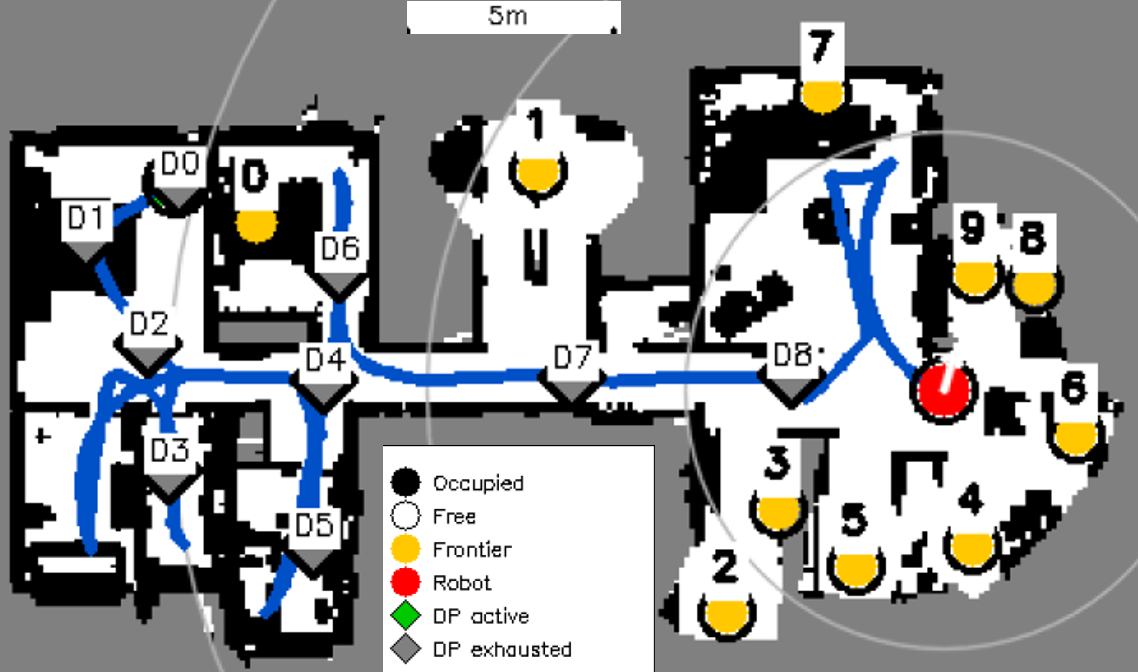}
\caption{Example \method{} decision point map slab after reachable-vicinity filtering. Occupied cells are shown in black, free space in white, active and exhausted decision points in green and gray, the recent trace in blue, the robot pose in red, and candidate frontiers in yellow.}
\label{fig:horus_decision_point_example}
\end{figure}

At ambiguous branch states, we pause the drone in place and execute a $360^\circ$ yaw
rotation before committing to the next frontier. This pan refreshes the local branch evidence
without traversing distance on an early move. After the pan, we render a top-down local map slab from the current voxel map, as shown in Fig.~\ref{fig:horus_decision_point_example}, indicating known free
space, obstacles, the recent robot trace, the current pose, and the reachable frontier candidates.
During the in-place
$360^\circ$ pan, we also sample RGB frames from the live camera stream while the robot rotates
in yaw at hover. Each frontier is then matched to at most one RGB frame by comparing the
frontier bearing from the rotation pose against the yaw of every captured RGB frame and selecting the
smallest wrapped angular error. Matches whose yaw error exceeds a threshold of $45^\circ$ are discarded,
so the selector sees only directionally aligned frontier imagery. As seen in Figure~\ref{fig:horus_falcon_stack}, MapServer supplies the Map Slab context, while, the $360^\circ$ Pan and
Frontier-View Match blocks together provide the visual evidence as input to 
Local Selector.

\subsection{Local Selector Policies}

Given the same decision gate and selector interface, we can swap in different local selector policies. The no-model
control, \methodnfp{}, or ``Nearest Frontier + Pan,'' makes the final choice with a purely geometric rule. Concretely, it
selects the reachable frontier with the lowest shared planner cost over the filtered candidate set.
This distinguishes it from \nf{}, or ``Nearest Frontier,'' which uses the same
nearest-frontier selection rule without the added pan step and is studied later in the ablation
experiments. \methodv{}, a visual-language model variant, uses map slabs, frontier imagery, planner cost,
and frontier coordinates. \methodof{}, another visual-language model variant  uses frontier imagery together with planner cost and frontier
coordinates, but without map slabs. \methodl{} uses language-only metadata such as planner
cost and frontier coordinates, but without frontier images.

\begin{figure}[!t]
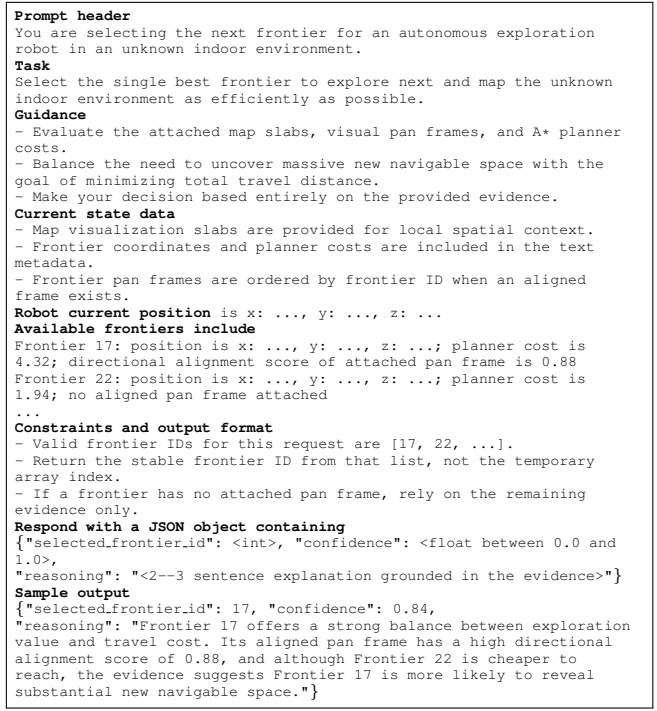

\centering
\setlength{\fboxsep}{3pt}
\fbox{%
\begin{minipage}{0.95\columnwidth}
\ttfamily
\fontsize{5.6}{6.2}\selectfont
\raggedright
\setlength{\parindent}{0pt}
\textbf{Prompt header}

You are selecting the next frontier for an autonomous exploration robot in an unknown indoor environment.

\textbf{Task}

Select the single best frontier to explore next and map the unknown indoor environment as efficiently as possible.

\textbf{Guidance}

- Evaluate the attached map slabs, visual pan frames, and A* planner costs.

- Balance the need to uncover massive new navigable space with the goal of minimizing total travel distance.

- Make your decision based entirely on the provided evidence.

\textbf{Current state data}

- Map visualization slabs are provided for local spatial context.

- Frontier coordinates and planner costs are included in the text metadata.

- Frontier pan frames are ordered by frontier ID when an aligned frame exists.

\textbf{Robot current position} is x: ..., y: ..., z: ...

\textbf{Available frontiers include}

Frontier 17: position is x: ..., y: ..., z: ...; planner cost is 4.32; directional alignment score of attached pan frame is 0.88

Frontier 22: position is x: ..., y: ..., z: ...; planner cost is 1.94; no aligned pan frame attached

\ldots

\textbf{Constraints and output format}

- Valid frontier IDs for this request are [17, 22, ...].

- Return the stable frontier ID from that list, not the temporary array index.

- If a frontier has no attached pan frame, rely on the remaining evidence only.

\textbf{Respond with a JSON object containing}

\texttt{\{"selected\_frontier\_id": <int>, "confidence": <float between 0.0 and 1.0>,}

\texttt{"reasoning": "<2--3 sentence explanation grounded in the evidence>"\}}

\textbf{Sample output}

\texttt{\{"selected\_frontier\_id": 17, "confidence": 0.84,}

\texttt{"reasoning": "Frontier 17 offers a strong balance between exploration value and travel cost. Its aligned pan frame has a high directional alignment score of 0.88, and although Frontier 22 is cheaper to reach, the evidence suggests Frontier 17 is more likely to reveal substantial new navigable space."\}}
\end{minipage}
}
\caption{\methodv{} prompt template and sample JSON response.}
\label{fig:horus_prompt_template}
\end{figure}

A representative prompt template for \methodv{}, together with a sample JSON response, is shown in
Fig.~\ref{fig:horus_prompt_template}. In all language-model variants, these local selection
requests are sent to the Gemini-2.0-Flash model. The prompt in the figure is intentionally
thoroughly structured to clearly define the robot's role, the decision objective, the evidence to
consider, and the required output format, while the sample response illustrates the expected
return schema and style of evidence-grounded justification. This helps keep the model's local
frontier selections consistent across branch decisions.

\subsection{Selection Tree Memory and Recovery Logic}

Local branch choices need to accumulate into a coherent long-horizon exploration strategy.
Conceptually, this part of \method{} is inspired by classical depth-first search with
backtracking over a search tree \cite{tarjan1972dfs}, as well as exploration methods that retain
an explicit tree of sensing or frontier decisions to support forward progress and later recovery
\cite{freda2008set,korb2018frontiertrees}. The robot commits to one local continuation, but
preserves enough state to later return to unresolved sibling branches once the current branch has
been exhausted.
Accordingly, when the reachable-vicinity set is empty, i.e., $|V_{r_v,t}|=0$, the robot invokes
this memory-and-recovery mechanism to revisit earlier unresolved branch alternatives.
To support that, we maintain lightweight tree-structured branch memory so the robot can
later revisit unresolved alternatives or relay through earlier branch states when a local region
has been exhausted.

Each ambiguous decision point records the available frontier IDs, the chosen continuation, and the
parent-child relation to earlier branch states. This yields a compact search tree over local
frontier decisions. When local progress is exhausted, the robot can backtrack to the nearest
non-exhausted ancestor, in the same spirit as DFS-style exploration over a branching graph
\cite{tarjan1972dfs,freda2008set,korb2018frontiertrees}. As
shown in Figure~\ref{fig:horus_falcon_stack}, this functionality is represented by the Selection
Tree Memory and Recovery Logic modules, which store the branch-state bookkeeping accumulated
during local decisions.

\section{Experimental Setup and Results}
\label{sec:eval}

In this section, we characterize the performance of OPAL and its variants in both simulation and actual hardware experiments.

\subsection{Benchmark Setup and Metrics}

\begin{figure}[!tbp]
\centering
\setlength{\tabcolsep}{4pt}
\begin{tabular}{cc}
  \includegraphics[width=0.46\linewidth]{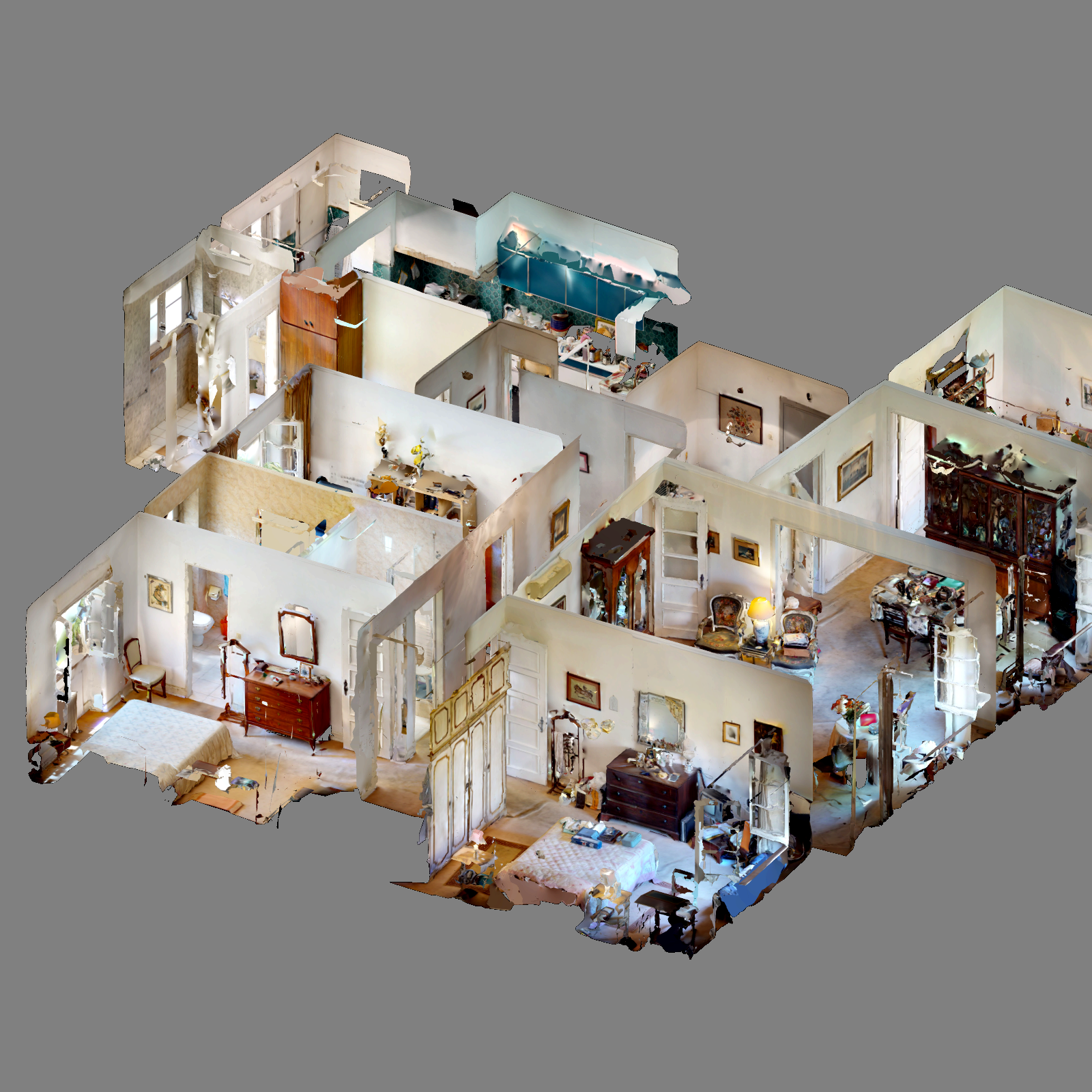} &
  \includegraphics[width=0.46\linewidth]{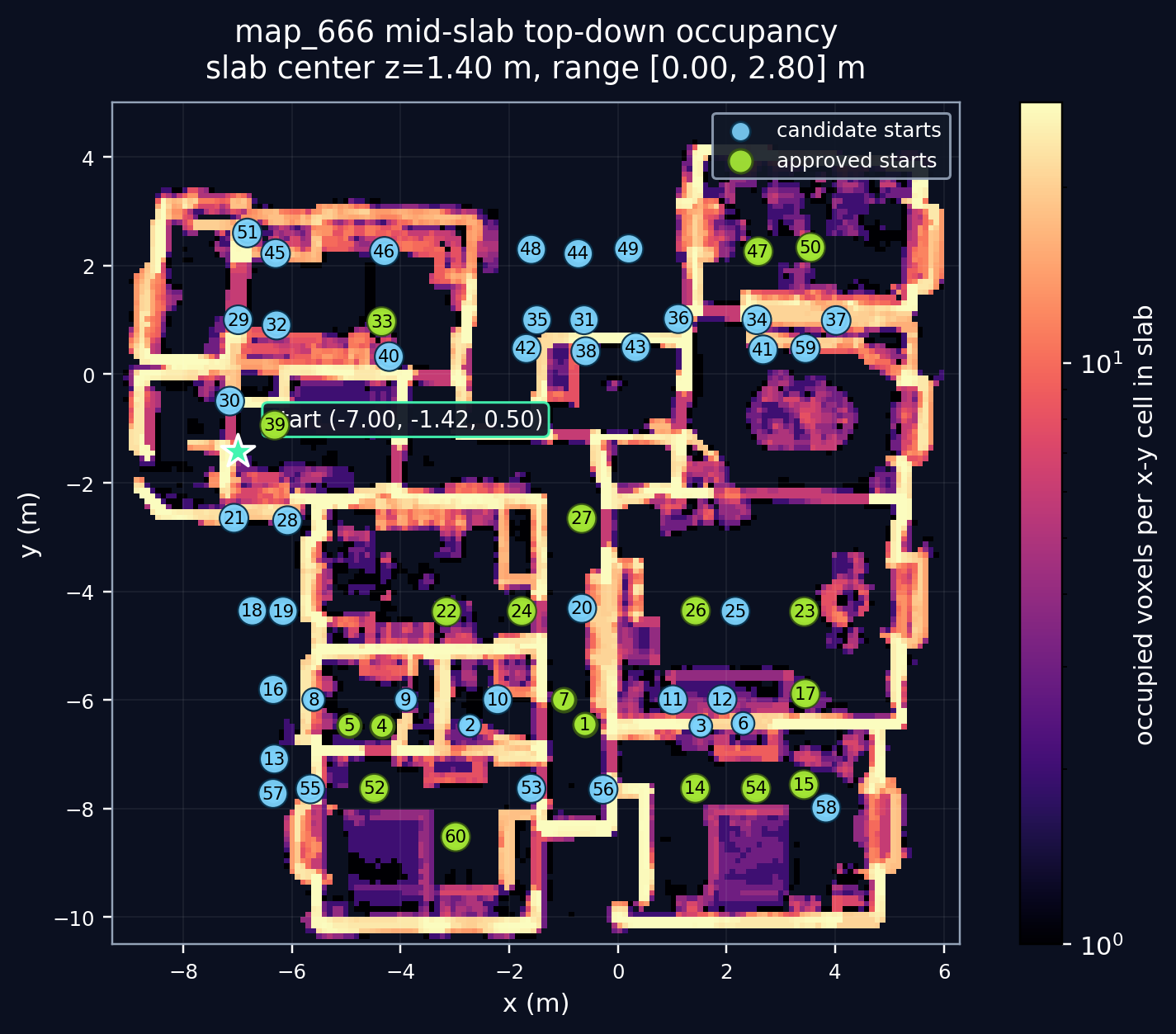} \\
  \scriptsize (a) &
  \scriptsize (b) \\
\end{tabular}
\caption{Map 666 from the HM3D benchmark. (a) The raw HM3D scene in dollhouse view before
aerial-simulation preprocessing, highlighting the multi-room layout, holes in the walls, and
internal cavities that we patch and seal to make the scene watertight. (b) Random-start selection
for Map 666. We first generate many random candidate starts shown in blue, then manually keep only the
points that lie in free space rather than inside walls; the valid starts used for evaluation are shown in green.}
\label{fig:hm3d_dollhouse_views}
\end{figure}

We use six Habitat-Matterport 3D (HM3D) \cite{ramakrishnan2021hm3d} indoor maps, namely 130, 195, 231, 417, 666, and 804
in our simulation trials. These maps are sufficiently large and topologically varied that
hallway and doorway branch choices materially affect the remainder of the trajectory. They also have RGB-textured meshes, which are important for collecting frontier images for VLM variants.
The benchmark maps used by FALCON \cite{zhang2024falcon} and EDEN \cite{dong2025eden} are largely plain white scenes and therefore much less
informative. Before evaluation, we use Blender \cite{blender} to preprocess the HM3D
meshes by manually patching holes in the walls and sealing internal cavities to make the scenes
watertight. To evaluate methods under diverse conditions, we generate random candidate start
points for each map and manually discard those that fall inside walls until 25 valid start poses remain. These valid starts are
used for 50 evaluation runs. For a given map and run index, all methods use the same
start position. Figure~\ref{fig:hm3d_dollhouse_views}
shows one benchmark scene in dollhouse view together with the valid random-start distribution
used for evaluation on Map 666, with the raw mesh shown in Fig.~\ref{fig:hm3d_dollhouse_views}(a) and the
filtered start set shown in Fig.~\ref{fig:hm3d_dollhouse_views}(b).

We compare \method{} variants and the external
baselines EDEN \cite{dong2025eden} and FALCON \cite{zhang2024falcon}. In simulation and later in hardware experiments, all methods use a depth sensor of 5\,meters range. We report performance using total traversed distance and coverage-versus-distance
area under the curve (AUC). The use of AUC is inspired by recent
active mapping and next-best-view work that evaluates coverage accumulation over distance via coverage
curves. Specifically, GLEAM \cite{chen2025gleam} measures AUC of cumulative coverage over exploration duration,
GenNBV \cite{chen2024gennbv} uses mean AUC of coverage ratio as a principal reconstruction metric, and
Hestia \cite{lu2026hestia} reports gains in the area under the coverage ratio curve.
We adapt AUC to our exploration setting by using traversed distance as the independent variable, so the metric reflects how efficiently a method converts motion into coverage.
While the total traversed distance summarizes when a run completes accumulating new coverage, AUC measures how much
coverage is accumulated throughout the entire exploration. For map $m$, method $a$, and
run $r$, let $C_{m,a,r}(d)$ denote the observed coverage after traveling distance $d$, and let
$d^{\mathrm{term}}_{m,a,r}$ denote the terminal distance of that run. Because different runs on the
same map can terminate after different traveled distances, we evaluate every run over a common
integration interval to compare AUC across runs and methods. Specifically, for each map $m$, we define
\begin{equation}
d_m^{\max} = \max_{a,r} d^{\mathrm{term}}_{m,a,r}.
\end{equation}
We then define the extrapolated coverage function
\begin{equation}
\widetilde{C}_{m,a,r}(d)=
\begin{cases}
C_{m,a,r}(d), & 0 \le d \le d^{\mathrm{term}}_{m,a,r}, \\
C_{m,a,r}\!\left(d^{\mathrm{term}}_{m,a,r}\right), & d^{\mathrm{term}}_{m,a,r} < d \le d_m^{\max}.
\end{cases}
\end{equation}
That is, once run $(m,a,r)$ reaches its terminal distance, its final observed coverage is held
constant for the remaining distance interval up to $d_m^{\max}$. The resulting AUC is
\begin{equation}
\mathrm{AUC}_{m,a,r} = \int_0^{d_m^{\max}} \widetilde{C}_{m,a,r}(d)\,\mathrm{d}d.
\end{equation}
Thus, $\mathrm{AUC}_{m,a,r}$ is the area under that run's coverage-versus-distance curve, measured over the
same distance range for every method and every run on the same map. Higher AUC therefore means the robot
covers more of the environment earlier in the trajectory.

\subsection{Language-Model Variants Versus No-Model OPAL}

\begin{figure}[!tbp]
\centering
\setlength{\tabcolsep}{2pt}
\begin{tabular}{c}
  \includegraphics[width=0.92\linewidth]{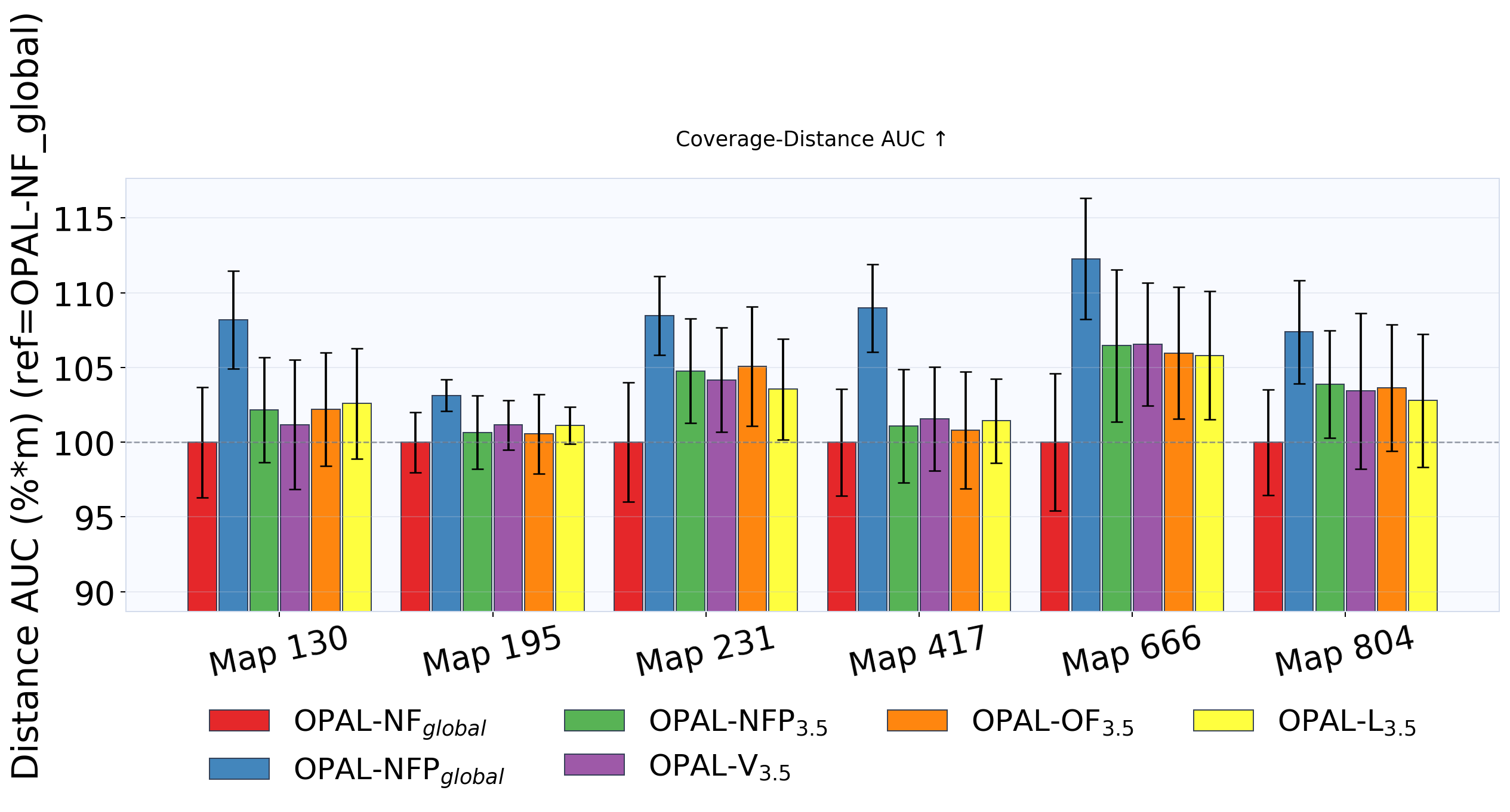} \\
  \scriptsize (a) \\[2pt]
  \includegraphics[width=0.92\linewidth]{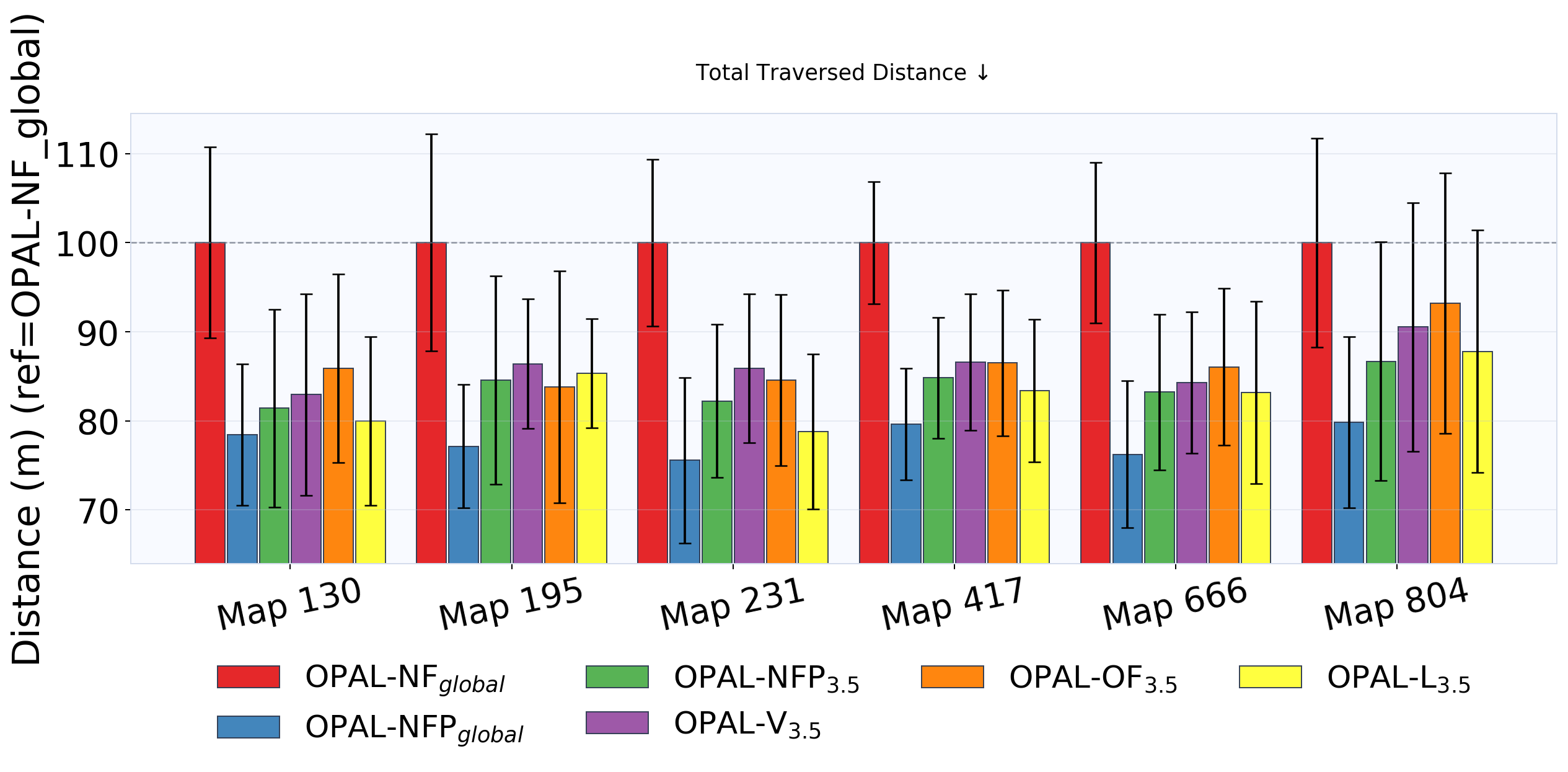} \\
  \scriptsize (b) \\[2pt]
  \includegraphics[width=0.92\linewidth]{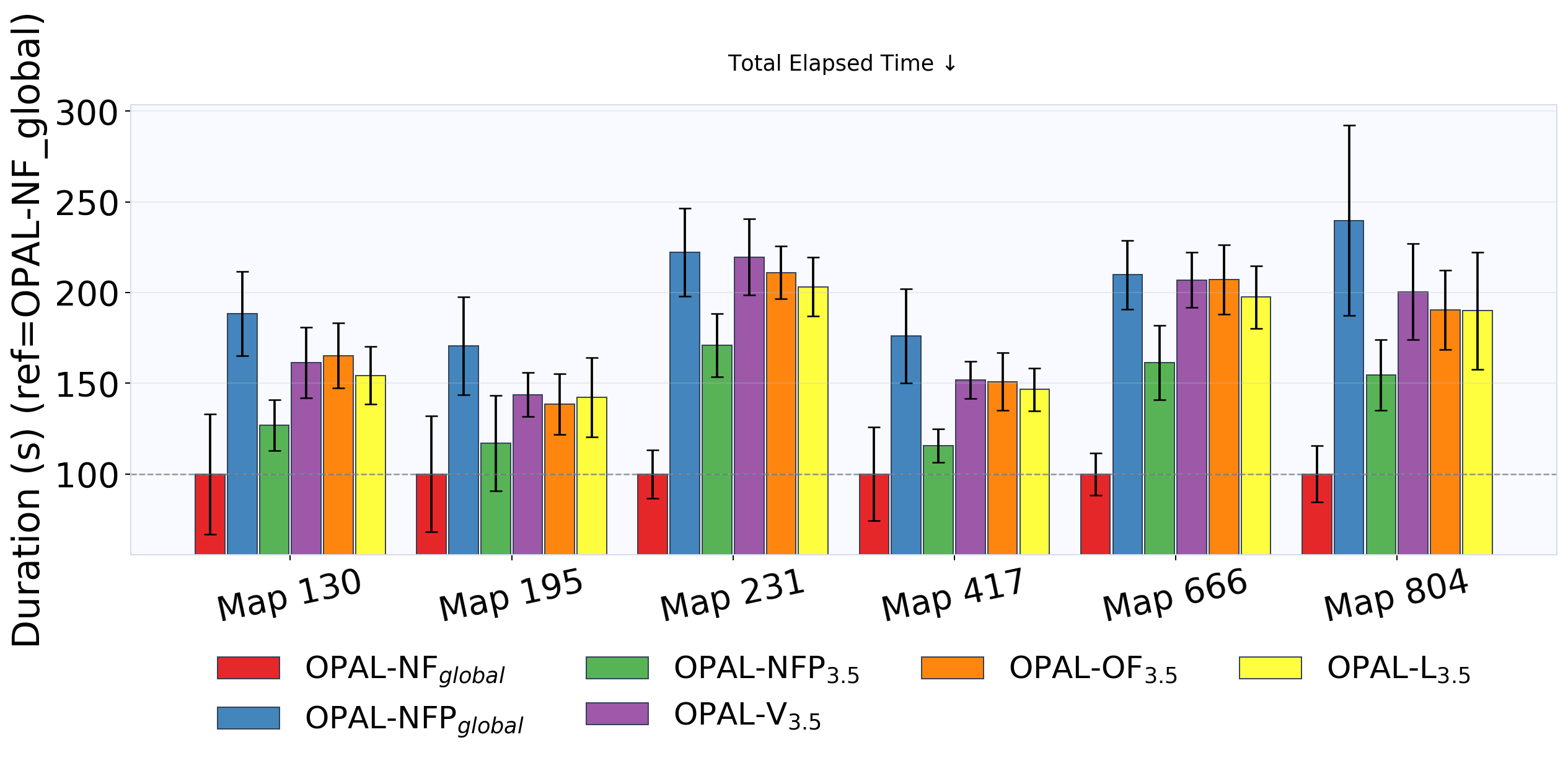} \\
  \scriptsize (c) \\[2pt]
  \includegraphics[width=0.92\linewidth]{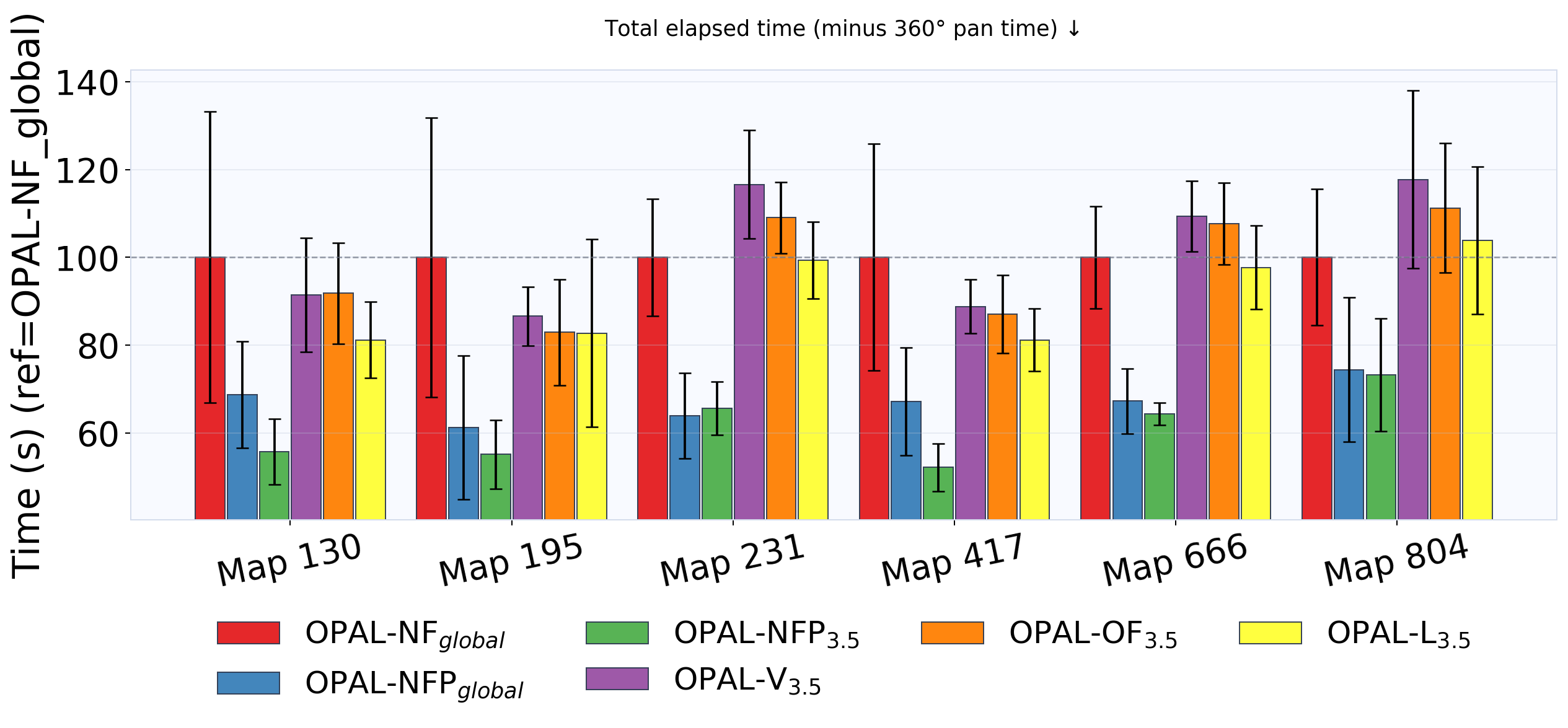} \\
  \scriptsize (d) \\
\end{tabular}
\caption{OPAL selector and geometric-policy comparison.
(a) Coverage-distance AUC. (b) Total traversed distance to full coverage, with reference values ranging from 138\,m to 320\,m.
(c) Average elapsed duration, with reference values ranging from 151\,s to 429\,s. (d) Duration after removing $360^\circ$
pan overhead.}
\label{fig:single_lm_comparison}
\end{figure}

We first examine whether language-model guidance improves exploration when the OPAL
decision scaffold is held fixed. In this comparison, all variants use
the same reachable-vicinity setting, $r_v=3.5$\,m. The model-based
variants are matched at this radius, and OPAL-NF$_{\mathrm{global}}$ and
OPAL-NFP$_{\mathrm{global}}$ are included as geometric reference policies.
Figure~\ref{fig:single_lm_comparison} summarizes the corresponding AUC,
traversed-distance, and timing trends, reported relative to the baseline OPAL-NF$_{\mathrm{global}}$.

Figure~\ref{fig:single_lm_comparison}(a) shows the coverage-versus-distance area under the curve for several OPAL variants.
Distance AUC is similar across OPAL-NFP$_{3.5}$, OPAL-OF$_{3.5}$,
OPAL-L$_{3.5}$, and OPAL-V$_{3.5}$, with all language-model variants tightly clustered around the no-model
OPAL-NFP$_{3.5}$ and differences that are small relative to the
standard deviation. A clear contrast appears between the global geometric
variants. OPAL-NFP$_{\mathrm{global}}$
achieves the highest AUC on every map, whereas OPAL-NF$_{\mathrm{global}}$ is
the lowest. This suggests that the main benefit in OPAL comes from the
structural policy choice, especially branch-point $360^\circ$ pan refresh and relaxed
locality, rather than from the specific language-model selector placed inside
the shared OPAL scaffold.

Figure~\ref{fig:single_lm_comparison}(b) shows the same pattern in total
traversed distance. At the common $r_v=3.5$\,m setting, OPAL-NFP and the
language-model variants remain close, with none of the language-model variants showing a
clear and consistent distance-efficiency advantage over the simpler no-model
baseline. The same contrast appears again between the global geometric
variants. OPAL-NFP$_{\mathrm{global}}$ has
the shortest traversed distance across maps, whereas
OPAL-NF$_{\mathrm{global}}$ has the longest one. In that
sense, OPAL-NFP$_{\mathrm{global}}$ is the best-performing policy in the figure when
total travel distance is the main objective.

Figure~\ref{fig:single_lm_comparison}(c) shows the elapsed time across OPAL variants for six maps.
Although the language-model variants remain close to OPAL-NFP
on AUC and traversed distance, they take more time to complete the exploration. OPAL-NFP$_{3.5}$ is the fastest, while the model-based variants are
slower in a way that tracks the
amount of information processed at each decision point. OPAL-V carries the
heaviest per-decision input burden, followed by OPAL-OF and OPAL-L. Among the global geometric methods, OPAL-NFP$_{\mathrm{global}}$ has a longer duration than
both OPAL-NFP$_{3.5}$ and OPAL-NF$_{\mathrm{global}}$. This is because OPAL-NFP$_{\mathrm{global}}$ performs more rotations than either of those two methods.

This is confirmed in Figure~\ref{fig:single_lm_comparison}(d), which shows
elapsed time after removing the $360^\circ$ pan, that is, the branch-point pan overhead.
Once that pan time is removed, both no-model pan variants,
OPAL-NFP$_{3.5}$ and OPAL-NFP$_{\mathrm{global}}$, drop below
OPAL-NF$_{\mathrm{global}}$ on every map. Furthermore, OPAL-NFP$_{\mathrm{global}}$ takes less time
than all the model-based variants, such as OPAL-OF, OPAL-V, and OPAL-L. This indicates that the large elapsed-time
penalty of the pan-based geometric methods in Figure~\ref{fig:single_lm_comparison}(c)
comes primarily from the pan itself. The timing of model-based variants also improves after removing the pan duration, but they still usually remain above the
no-model pan-based methods. This can be attributed to the model query and associated processing costs.

Figure~\ref{fig:single_lm_comparison}(d) is more than a bookkeeping exercise. Specifically,
on current aerial hardware, an in-place $360^\circ$ yaw sweep can take longer
than translating linearly through free space, but that penalty could be
reduced by mounting the RGB-D sensor on a relatively inexpensive dedicated spinning mechanism rather
than rotating the entire vehicle. In that hardware regime, the effective runtime
of OPAL could drop substantially relative to the totals in
Figure~\ref{fig:single_lm_comparison}(c).

From Figure~\ref{fig:single_lm_comparison} we conclude that across the AUC, traversed-distance, and timing results, the model-based
selectors do not result in substanstial exploration benefit to justify their added
query cost. Therefore, we compare the geometric variant of OPAL, namely OPAL-NFP,
as the representative OPAL variant for the reachable-vicinity radius study and for comparisons
against baselines such as FALCON and EDEN in the next section.

\subsection{Reachable-Vicinity Radius Sweep}

\begin{figure}[!tbp]
\centering
\setlength{\tabcolsep}{3pt}
\begin{tabular}{c}
  \includegraphics[width=0.90\linewidth]{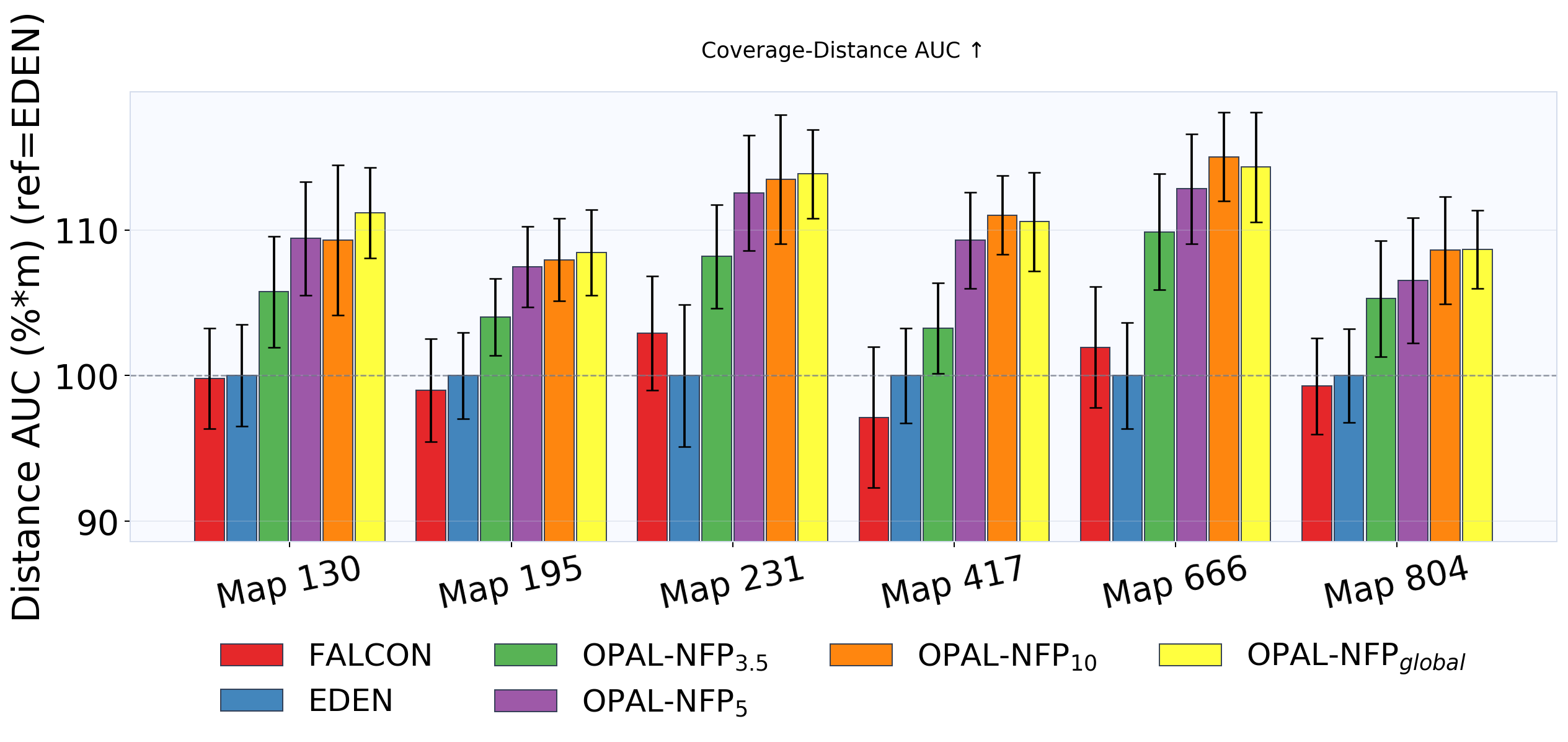} \\
  \scriptsize (a) \\[2pt]
  \includegraphics[width=0.90\linewidth]{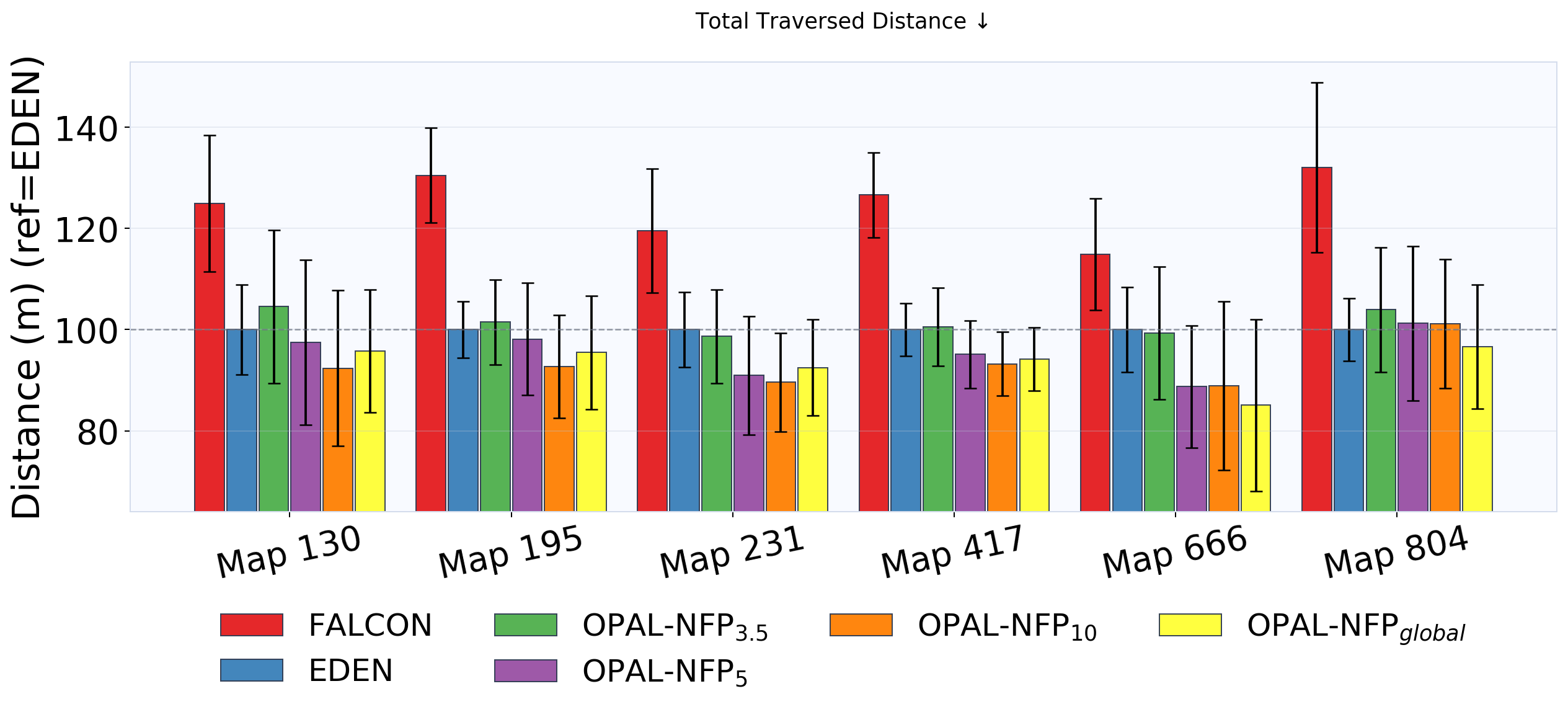} \\
  \scriptsize (b) \\[2pt]
  \includegraphics[width=0.90\linewidth]{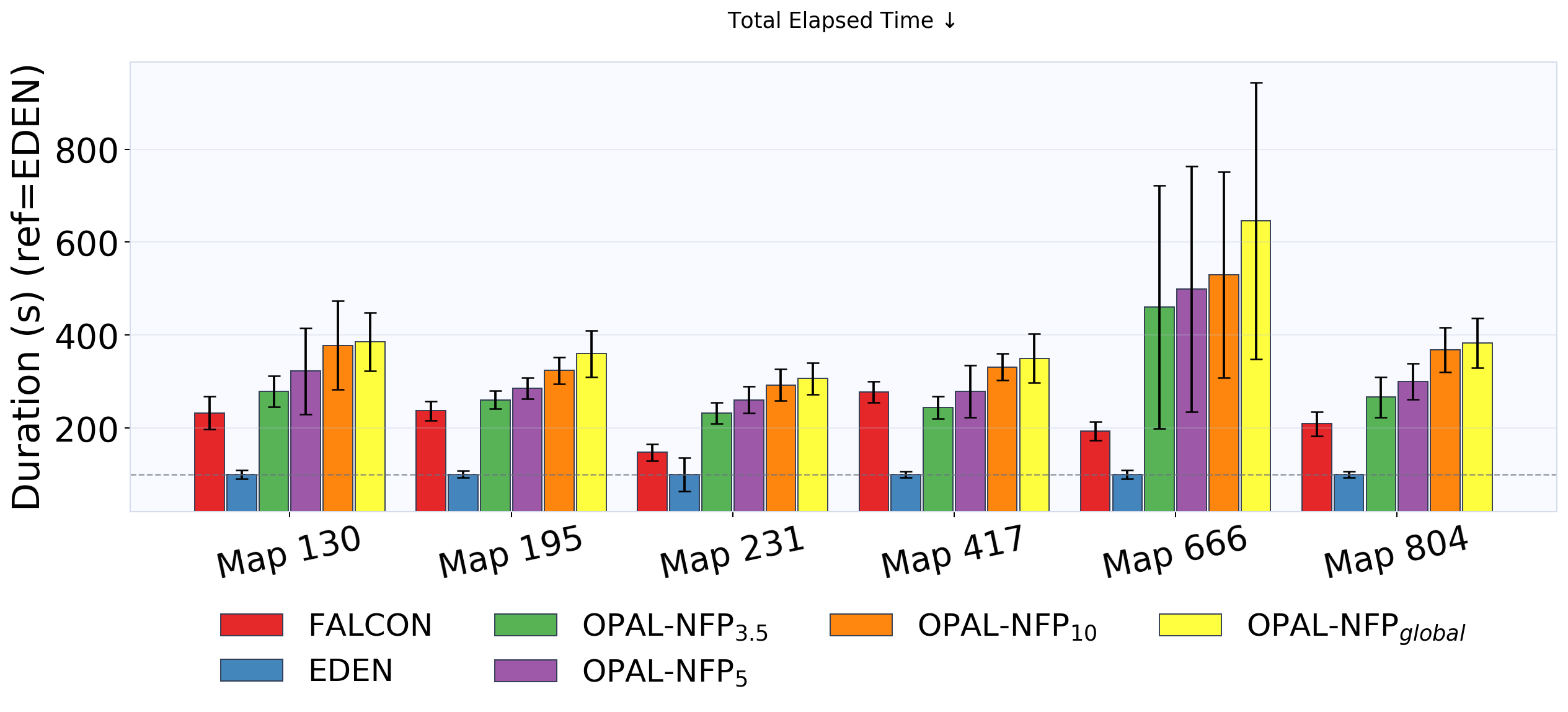} \\
  \scriptsize (c) \\[2pt]
  \includegraphics[width=0.90\linewidth]{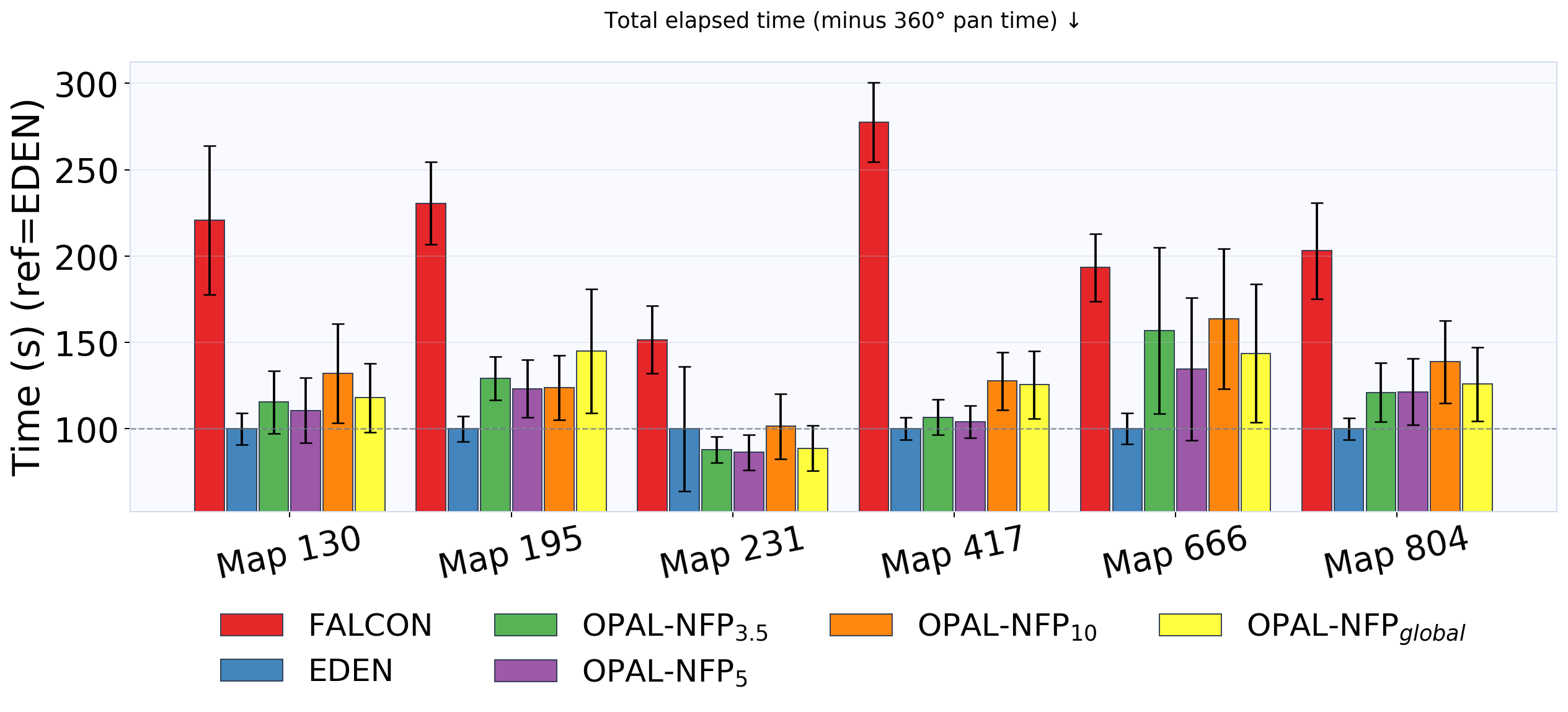} \\
  \scriptsize (d) \\[2pt]
  \includegraphics[width=0.90\linewidth]{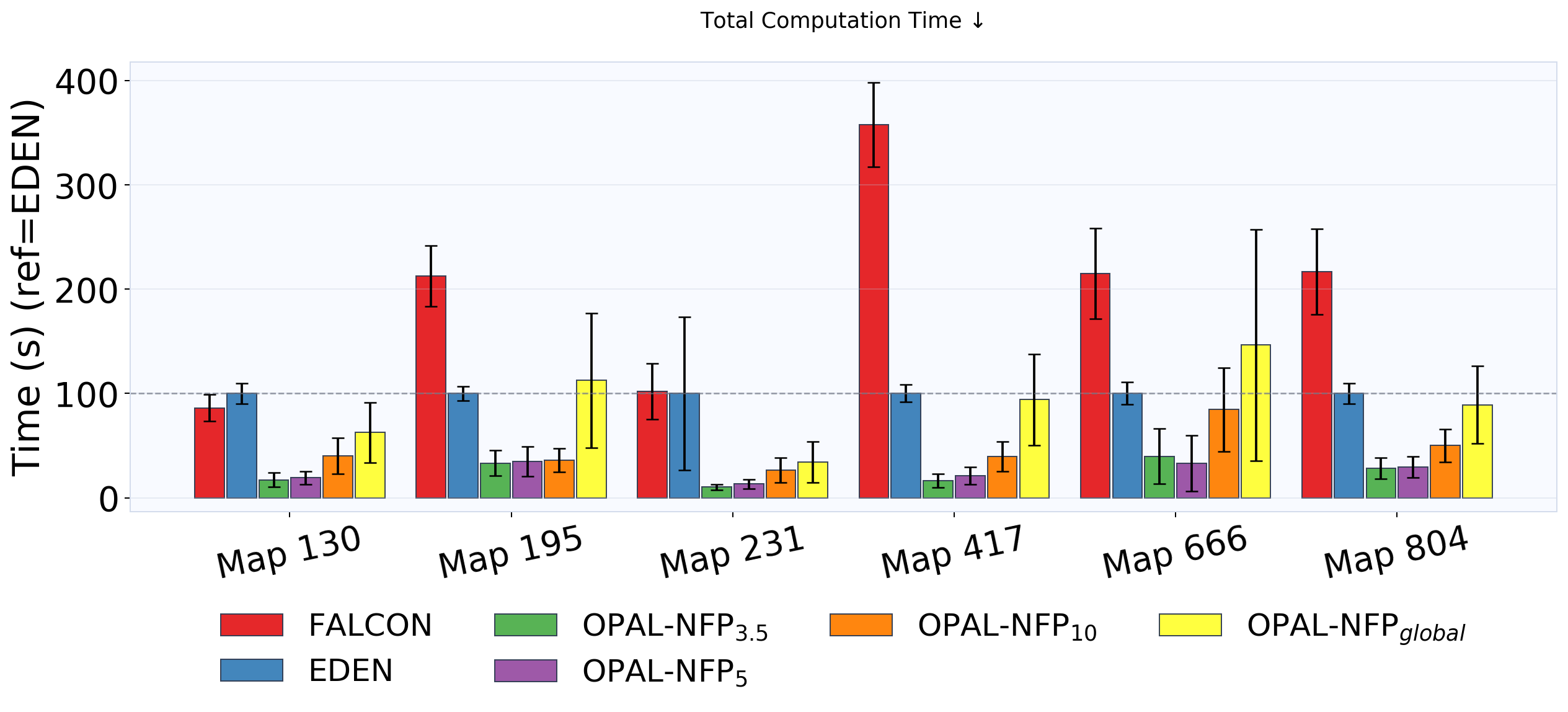} \\
  \scriptsize (e) \\
\end{tabular}
\caption{Reachable-vicinity radius $r_v$ sweep across the OPAL-NFP family and SoTA baselines.
(a) Coverage-distance AUC, where higher is better. (b) Total traversed distance
to full coverage, where lower is better, with reference values ranging from 115\,m to 266\,m. (c) Average elapsed duration by
map, where lower is better, with reference values ranging from 112\,s to 204\,s. (d) Elapsed duration after removing $360^\circ$ pan
time. (e) Full computation time, with reference values ranging from 31\,s to 79\,s.}
\label{fig:single_radius_sweep_main}
\end{figure}

We now investigate the radius $r_v$ choice in more detail by sweeping the
reachable-vicinity threshold within the OPAL-NFP family and comparing the resulting variants against EDEN and FALCON. The tested settings are
$r_v \in \{3.5, 5, 10\}$ meters, together with the fully global limit OPAL-NFP$_{\mathrm{global}}$. The corresponding trends are shown in
Figure~\ref{fig:single_radius_sweep_main}, with relative performance reported with respect to the baseline EDEN.

As shown in Figures~\ref{fig:single_radius_sweep_main}(a) and~\ref{fig:single_radius_sweep_main}(b), both coverage-distance AUC and total
traversed distance improve as the radius $r_v$ increases. The tight local setting,
OPAL-NFP$_{3.5}$, is consistently the worst performing member of the family on
exploration efficiency. Broader-radius policies recover coverage earlier and
reach high coverage with less travel. However, these gains do not continue
indefinitely. Across the benchmark, AUC and traversed distance largely plateau
by $r_v=10$ m, and OPAL-NFP$_{\mathrm{global}}$ adds only a small, map-dependent
change beyond that point.

As shown in Figure~\ref{fig:single_radius_sweep_main}(c), elapsed time follows
a different pattern. Unlike AUC and traversed distance, total elapsed time
increases steadily with radius. EDEN is the fastest method overall in this
panel, and within the OPAL-NFP family the tight local setting,
OPAL-NFP$_{3.5}$, is consistently the fastest, with elapsed time increasing with radius across all OPAL-NFP variants. This trend follows directly
from the decision rule. As $r_v$ grows, the reachable-vicinity set
$V_{r_v,t}$ is more likely to satisfy $|V_{r_v,t}|>1$, so the robot
encounters more ambiguous branch states and triggers more in-place
$360^\circ$ pans before committing to the next move.

Figure~\ref{fig:single_radius_sweep_main}(d) clarifies where much of that elapsed time
is spent by subtracting pan time. Once this overhead is removed, the ordering among methods
changes noticeably. Specifically, OPAL-NFP$_{3.5}$ and OPAL-NFP$_{5}$ become the fastest 
across most maps, while the larger-radius settings,
OPAL-NFP$_{10}$ and OPAL-NFP$_{\mathrm{global}}$, still remain slower. This
shows that the results in Figure~\ref{fig:single_radius_sweep_main}(c) are primarily driven by pan overhead.
Accordingly, Figure~\ref{fig:single_radius_sweep_main}(d) is a useful proxy for
hardware regimes where panoramic sensing is cheaper, for example, by mounting the depth camera on a motorized rotating mechanism.
In this case, the total elapsed time for OPAL settings would become more competitive with that of EDEN.

\subsection{Computation Time Analysis}

To separate runtime from the time spent executing physical motion, we
decompose elapsed time using the exploration finite-state machine (FSM). Let
$T_{\mathrm{elapsed}}$ denote total runtime, $T_{\mathrm{exec}}$ the time spent
in the trajectory-execution state, and $T_{\mathrm{pan}}$ the time spent
performing $360^\circ$ pan at decision points. We define
\begin{equation}
\begin{aligned}
T_{\mathrm{movement}} &= T_{\mathrm{exec}} + T_{\mathrm{pan}}, \\
T_{\mathrm{compute}} &= T_{\mathrm{elapsed}} - T_{\mathrm{movement}}.
\end{aligned}
\end{equation}
Since the FSM occupies only one state at a time, there is no double counting.

Figure~\ref{fig:single_radius_sweep_main}(e) shows the computation component
of all methods. Within the OPAL-NFP family, computation time increases with the
vicinity radius $r_v$, mirroring the increase in total elapsed time. This creates
a clear tradeoff in which larger radii improve coverage-distance AUC and reduce traversed distance,
whereas smaller radii yield lower duration and computation time.

As seen in Figure~\ref{fig:single_radius_sweep_main}(e), excluding OPAL-NFP$_{\mathrm{global}}$, every OPAL-NFP variant has
lower computation time than EDEN and FALCON. OPAL-NFP$_{10}$ reduces
computation time by 22.0\% to 68.1\% relative to OPAL-NFP$_{\mathrm{global}}$
while maintaining similar performance in coverage-distance AUC, traversed distance, and
elapsed duration. For this reason, we use OPAL-NFP$_{10}$ in the hardware experiments in Section~\ref{subsec:hard_val}.

Table~\ref{tab:eden_vs_nfp10_vic} compares EDEN and OPAL-NFP$_{10}$ across multiple metrics for six maps.
As shown, OPAL-NFP$_{10}$ outperforms EDEN on nearly all maps in coverage-distance AUC, total traversed distance,
and computation time. Averaged across maps, OPAL-NFP$_{10}$ achieves 10.9\% higher coverage-distance AUC,
7.0\% shorter traversed distance, and 53.8\% lower computation time than EDEN. However, EDEN is, on average, 3.7 times faster in elapsed time than OPAL-NFP$_{10}$.
We expect similar trends to hold for other $r_v$ values of OPAL-NFP.

\begin{table}[t]
\centering
\tiny
\setlength{\tabcolsep}{3pt}
\renewcommand{\arraystretch}{1.08}
\resizebox{\columnwidth}{!}{%
\begin{tabular}{llccccccc}
\hline
Metric & Quantity & Map 130 & Map 195 & Map 231 & Map 417 & Map 666 & Map 804 & Average \\
\hline
Coverage-Distance AUC $\uparrow$ & EDEN & 22362.4 & 34386.5 & 12157.9 & 24430.0 & 14793.7 & 18030.9 & \\
 & \method{}-NFP$_{10}$ & \textbf{24447.1} & \textbf{37124.3} & \textbf{13797.1} & \textbf{27124.8} & \textbf{17017.5} & \textbf{19583.8} & \\
 & OPAL gain/loss w.r.t. EDEN & +9.3\% & +8.0\% & +13.5\% & +11.0\% & +15.0\% & +8.6\% & +10.9\% \\
\hline
Distance (m) $\downarrow$ & EDEN & 164.9 & 266.9 & 115.2 & 224.2 & 141.2 & \textbf{163.5} & \\
 & \method{}-NFP$_{10}$ & \textbf{152.3} & \textbf{247.4} & \textbf{103.2} & \textbf{209.1} & \textbf{125.5} & 165.4 & \\
 & OPAL gain/loss w.r.t. EDEN & -7.6\% & -7.3\% & -10.4\% & -6.7\% & -11.1\% & +1.2\% & -7.0\% \\
\hline
Computation (s) $\downarrow$ & EDEN & 51.0 & 79.2 & 49.2 & 72.6 & 31.9 & 36.7 & \\
 & \method{}-NFP$_{10}$ & \textbf{20.6} & \textbf{28.4} & \textbf{13.1} & \textbf{28.7} & \textbf{27.0} & \textbf{18.4} & \\
 & OPAL gain/loss w.r.t. EDEN & -59.6\% & -64.1\% & -73.4\% & -60.4\% & -15.3\% & -49.9\% & -53.8\% \\
\hline
Time (s) $\downarrow$ & EDEN & \textbf{134.1} & \textbf{204.9} & \textbf{113.3} & \textbf{182.9} & \textbf{112.4} & \textbf{126.4} & \\
 & \method{}-NFP$_{10}$ & 506.9 & 663.1 & 331.6 & 605.4 & 595.7 & 465.8 & \\
 & OPAL gain/loss w.r.t. EDEN & 3.8x & 3.2x & 2.9x & 3.3x & 5.3x & 3.7x & 3.7x \\
\hline
\end{tabular}%
}
\caption{Per-map comparison of EDEN and \method{}-NFP$_{10}$}
\label{tab:eden_vs_nfp10_vic}
\end{table}

\subsection{Hardware Validation}
\label{subsec:hard_val}

\begin{figure}[!tbp]
\centering
\setlength{\tabcolsep}{3pt}
\begin{tabular}{cc}
  \includegraphics[width=0.47\linewidth]{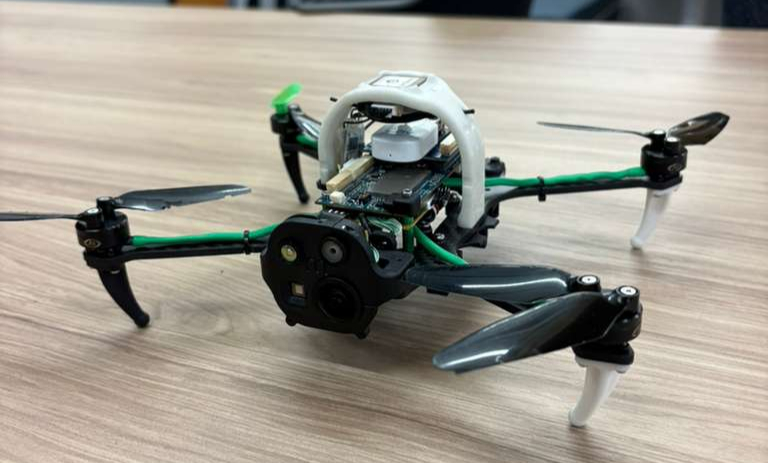} &
  \includegraphics[width=0.47\linewidth]{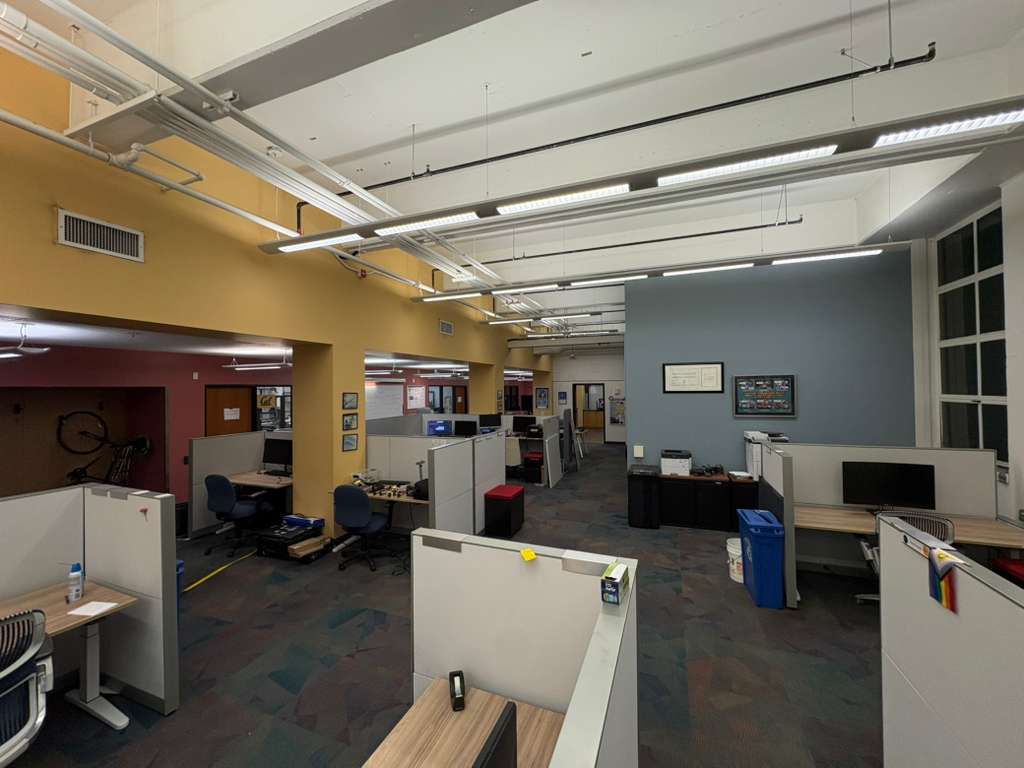} \\
  \scriptsize (a) Starling 2 hardware platform &
  \scriptsize (b) Forward RGB camera observation \\
\end{tabular}
\caption{Hardware-validation setup. (a) The ModalAI Starling~2 platform used
for the indoor flight tests. (b) Representative forward RGB observation}
\label{fig:single_hardware_setup}
\end{figure}

We use the ModalAI Starling~2 \texttt{C27} platform for actual hardware experiments. It is built
around a VOXL~2 board, with three AR0144 tracking cameras for VIO, one IMX412 RGB camera
for observation capture, and one LIOW2 ToF sensor for depth \cite{modalai_starling2_d0014,modalai_voxl2_coax_bundles,modalai_openvins_server}.
For hardware experiments as well, we limit the depth sensor range to 5\,m.
Modal Pipe Architecture (MPA) services \cite{modalai_mpa} bridge the relevant
sensor and estimator streams into ROS, while the planner publishes B-spline
trajectories that are converted into ModalAI \texttt{trajectory\_t} messages and
handed to \texttt{voxl-vision-hub} for offboard execution on PX4 \cite{modalai_voxl_vision_hub,modalai_voxl2_px4}. This keeps the exploration-side
mapper and planner observable in ROS while leaving the flight-critical odometry
and trajectory handoff inside the native VOXL-to-PX4 stack.
Figure~\ref{fig:single_hardware_setup} shows the hardware platform and a
representative onboard RGB observation.

We restrict the hardware baseline comparison to FALCON. EDEN's released system
is coupled to a different end-to-end sensing, mapping, and execution stack
than the Starling~2 platform used here, so reproducing it onboard would have required a substantial platform-specific reimplementation rather than a
planner-only swap.

Table~\ref{tab:hardware_validation_time_distance} summarizes the
coverage-distance AUC, traversed distance, and total elapsed time measured during hardware validation runs in two indoor layouts.
In Layout 1, we compare FALCON, OPAL-V$_{3.5}$ and OPAL-NFP$_{3.5}$. As seen, for Layout 1, OPAL-NFP$_{3.5}$
improves coverage-distance AUC by 10.1\% and reduces traversed distance by 15.8\%
relative to FALCON. In Layout 2, OPAL-NFP$_{10}$ has 25.5\% shorter distance than FALCON and 
7.8\% higher coverage-distance AUC than FALCON. In Layout 2, OPAL-NFP$_{10}$  also outperforms 
OPAL-NFP$_{3.5}$ in both metrics.
This is consistent with the simulation-side trend that relaxing the vicinity
radius improves distance-efficiency metrics. 
Although OPAL variants improve AUC and distance, FALCON is the 
fastest method in both layouts, followed by OPAL-NFP$_{3.5}$.

\begin{table}[!tbp]
\centering
\scriptsize
\setlength{\tabcolsep}{2pt}
\renewcommand{\arraystretch}{1.08}
\resizebox{\columnwidth}{!}{%
\begin{tabular}{llccc}
\hline
Layout & Method & \shortstack[c]{Coverage\\Distance\\AUC $\uparrow$} & \shortstack[c]{Distance (m) $\downarrow$} & \shortstack[c]{Total\\elapsed\\time (s) $\downarrow$} \\
\hline
1 & FALCON & 2405.6  & 26.0 & \textbf{36.7} \\
 & \methodv & 2544.8 & 25.5 & 132.7 \\
 & \methodnfpr{3.5} & \textbf{2647.4} & \textbf{21.9} & 56.0 \\
 & \methodnfpr{3.5} w.r.t.\ FALCON & +10.1\% & -15.8\% & +52.6\% \\
\hline
2 & FALCON & 2755.8 & 32.5 & \textbf{32.0} \\
 & \methodnfpr{3.5} & 2754.6 & 27.8 & 41.3 \\
 & \methodnfpr{10} & \textbf{2970.1} & \textbf{24.2} & 54.0 \\
 & \methodnfpr{10} w.r.t.\ FALCON & +7.8\% & -25.5\% & +68.8\% \\
\hline
\end{tabular}
}
\caption{Hardware-validation distance AUC, total elapsed time, and total traversed distance.}
\label{tab:hardware_validation_time_distance}
\end{table}

\begin{figure}[!tbp]
\centering
\setlength{\tabcolsep}{1pt}
\begin{tabular}{cc}
  \multicolumn{2}{c}{\scriptsize \methodnfpr{10}} \\[-1pt]
  \includegraphics[width=0.29\linewidth]{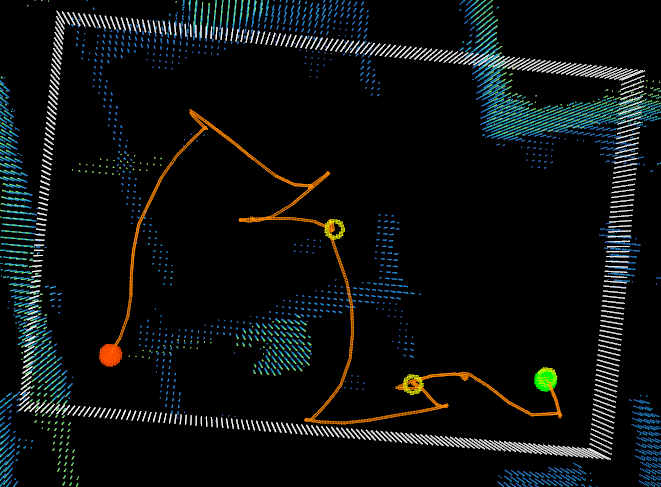} &
  \includegraphics[width=0.67\linewidth]{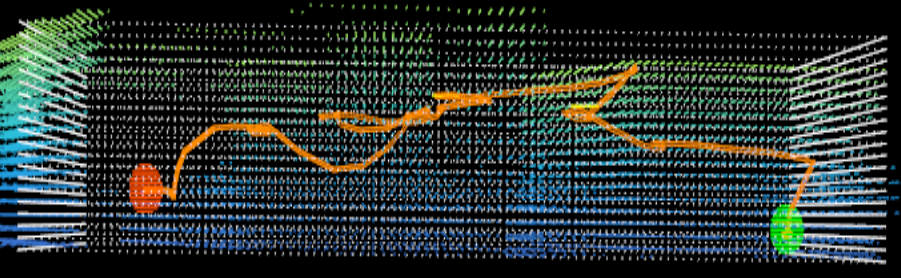} \\
  \scriptsize (a) &
  \scriptsize (d) \\[2pt]
  \multicolumn{2}{c}{\scriptsize \methodnfpr{3.5}} \\[-1pt]
  \includegraphics[width=0.29\linewidth]{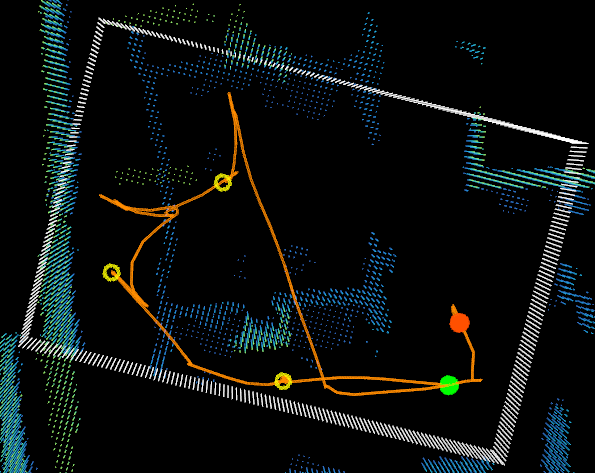} &
  \includegraphics[width=0.67\linewidth]{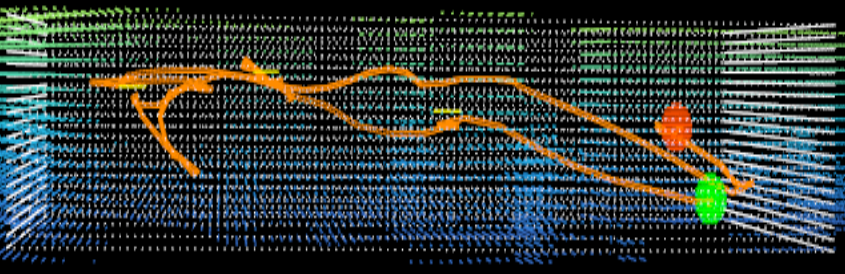} \\
  \scriptsize (b) &
  \scriptsize (e) \\[2pt]
  \multicolumn{2}{c}{\scriptsize FALCON} \\[-1pt]
  \includegraphics[width=0.29\linewidth]{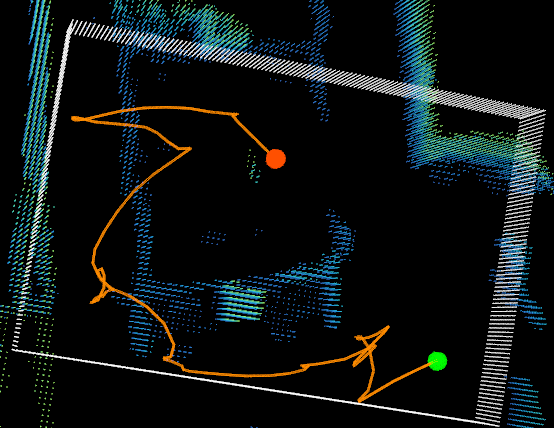} &
  \includegraphics[width=0.67\linewidth]{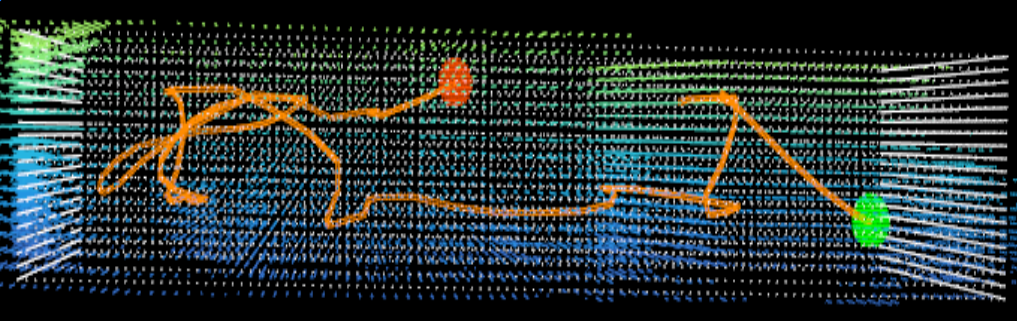} \\
  \scriptsize (c) &
  \scriptsize (f) \\
\end{tabular}
\caption{Representative voxel maps from hardware runs in Indoor Layout 2. Top down view for (a) OPAL-NFP$_{10}$, (b) OPAL-NFP$_{3.5}$ and (c) FALCON. Side views of (d) OPAL-NFP$_{10}$, (e) OPAL-NFP$_{3.5}$ and (f) FALCON. The x and y axes for the side views have different scales to aid with trajectory visualization. Voxel height is colored with a blue-to-green gradient, blue closer to the ground. The green circle marks the start point, the red one marks the end point, the orange trace shows the executed trajectory, yellow discs mark locations where the drone performs an in-place $360^\circ$ pan, and the white rectangle shows the exploration bounds.}
\label{fig:single_hardware_voxel_maps}
\end{figure}

Figure~\ref{fig:single_hardware_voxel_maps} shows representative traces from
the hardware comparisons in Layout 2. The visuals are in agreement 
with the travelled distance metric 
in Table~\ref{tab:hardware_validation_time_distance}, in that OPAL-NFP$_{10}$ has the shortest distance, 
followed by OPAL-NFP$_{3.5}$, with FALCON resulting in the longest path.

\section{Conclusions}
\label{sec:conclusion}

We introduce \methodfull{}, a frontier-exploration scaffold that adds a $360^\circ$ yaw pan at
branch points before selecting the next frontier. Across matched comparisons, a simple
geometric heuristic that selects the frontier within a sphere of radius $r_v$ with the
lowest local planner cost performs nearly as well as the LLM/VLM variants, indicating that
most of the gain comes from the $360^\circ$ yaw pan rather than from language-model
reasoning. The vicinity radius $r_v$ defines the main tradeoff. Smaller $r_v$ reduces
elapsed time and computation time but lowers AUC, whereas larger $r_v$ improves AUC and final
distance. Based on simulation results on six maps, we conclude that OPAL-NFP$_5$ and OPAL-NFP$_{10}$
outperform EDEN in coverage-distance AUC, total traversed distance, and computation time, whereas EDEN retains a lower elapsed time.
Hardware experiments further support these findings, showing that OPAL's distance-efficiency advantages carry
over to onboard flight tests. As for future work, it would be valuable to test a hardware modification that enables faster
panoramic sensing, such as mounting the depth camera on an inexpensive, lightweight, dedicated spinning mechanism.

% \clearpage
\bibliographystyle{IEEEtran}
\bibliography{IEEEabrv,references,paper_refs}

\end{document}